\documentclass[runningheads]{llncs}
\usepackage{graphicx}
\usepackage{comment}
\usepackage{amsmath,amssymb} 
\usepackage{color}
\usepackage{subcaption}
\usepackage{bbm}
\usepackage{booktabs}
\usepackage{algpseudocode}
\usepackage[ruled]{algorithm2e}
\usepackage{wrapfig}
\usepackage[mathscr]{euscript}
\let\euscr\mathscr \let\mathscr\relax
\usepackage[scr]{rsfso}
\usepackage{placeins}
\usepackage{graphics}
\usepackage{hyperref}
\usepackage[capitalise]{cleveref}
\begin{document}
\title{Mining Label Distribution Drift in Unsupervised Domain Adaptation}
%
%
\author{Peizhao Li\inst{1} \and Zhengming Ding\inst{2} \and Hongfu Liu\inst{1}}
\authorrunning{P. Li et al.}
%
\institute{Brandeis University \and Tulane University}
\maketitle              
\begin{abstract}
Unsupervised domain adaptation targets to transfer task-related knowledge from labeled source domain to unlabeled target domain. Although tremendous efforts have been made to minimize domain divergence, most existing methods only partially manage by aligning feature representations from diverse domains. Beyond the discrepancy in data distribution, the gap between source and target label distribution, recognized as label distribution drift, is another crucial factor raising domain divergence, and has been under insufficient exploration. From this perspective, we first reveal how label distribution drift brings negative influence. Next, we propose Label distribution Matching Domain Adversarial Network (LMDAN) to handle data distribution shift and label distribution drift jointly. In LMDAN, label distribution drift is addressed by a source sample weighting strategy, which selects samples that contribute to positive adaptation and avoid adverse effects brought by the mismatched samples. Experiments show that LMDAN delivers superior performance under considerable label distribution drift.

\keywords{Unsupervised Domain Adaptation \and Label Distribution Drift \and Transfer Learning \and Deep Learning.}
\end{abstract}
\section{Introduction}

\begin{figure*}[t]
\centering
\begin{subfigure}{0.48\columnwidth}
    \includegraphics[width=0.96\columnwidth]{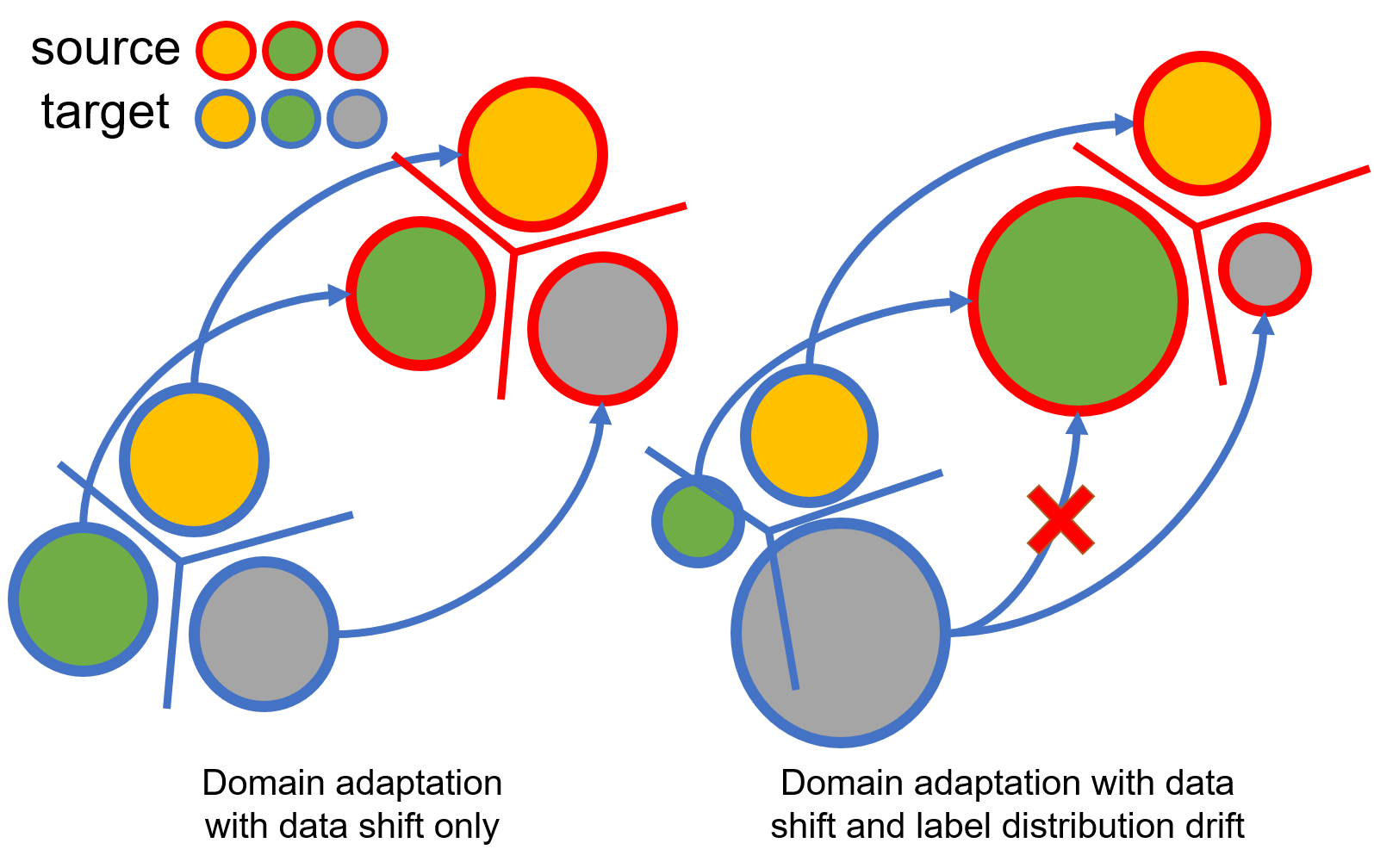}
    \caption{\scriptsize{Schematic illustration}}
    \label{fig:motivation1}
\end{subfigure}
\begin{subfigure}{0.48\columnwidth}
    \includegraphics[width=0.96\columnwidth]{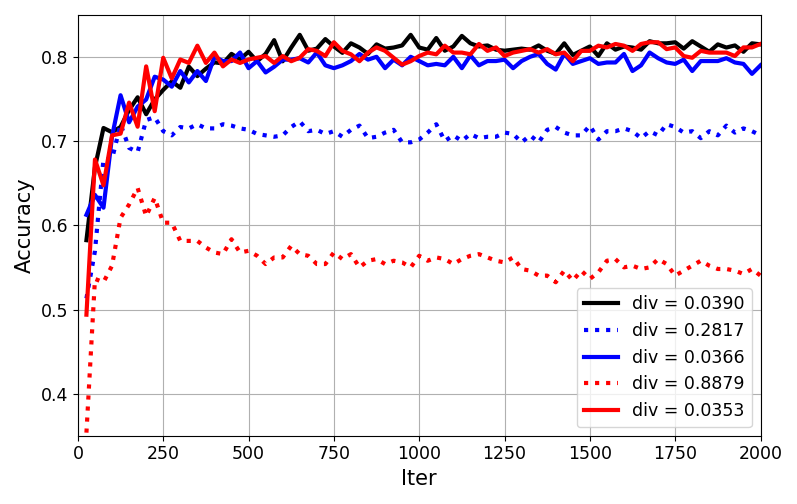}
    \caption{\scriptsize{Empirical illustration}}
    \label{fig:motivation2}
\end{subfigure}
\setlength{\belowcaptionskip}{-0.5cm}
\caption{(a) Schematic illustration on domain adaptation with and w/o label distribution drift. Source and target domain differ in the color of the borders of the circles. Circles with different colors inside denote different categories, and the size indicates the number of samples within that category. Straight lines denote the decision boundary learned by the classifier. Adaptation under label distribution drift makes features misaligned at the categorical level and decision boundary not applicable to target domain. (b) Empirical illustration on the performance of DANN under varying degrees of label distribution drift on \textit{Office-31} dataset from source domain \textit{Amazon} to target domain \textit{Webcam}. The black line indicates training with the original label distribution, while the blue and red lines denote sample drop rates at 50\% and 75\%, respectively. Solid lines indicate that dropped samples come from the first 15 classes for both source and target domain, while dashed lines indicate drops come from the first 15 and last 16 classes in source and target domain. The legend provides KL-divergence between source and target label distribution.}
\label{fig:motivation}
\end{figure*}

Domain adaptation is a fundamental research topic in the machine learning and computer vision field~\cite{ben2010,Ding2019}. It aims to build models on labeled source data and related target data, then make models adapt and generalize on target domain. Different settings for domain adaptation are applicable for complicated real-world problems~\cite{zeroshot,liumultisource,li2022exploiting,dingzeroshot,8362753,8398464,ding2017deep,li2021selfdoc,li2020deep}. Unsupervised domain adaptation, containing no label in target domain, is a challenging but practical setting, owing to that actual scenes are actually suffering from the lack of specific annotations.

The mitigation of domain shift, which aims to reduce the domain divergence between source and target, is the primary solution for unsupervised domain adaptation problems. Existing methods~\cite{GFK,dan,DANN,JAN,wmmd,cdan,sysnets,ding2018robust,Ding_2018_ECCV} mainly focus on alleviating negative influence brought by domain shift in feature representations. They reduce the discrepancy by pushing feature distribution from two separate domains close to each other. Consequently, models are expected to be generalized favorably to a related target data distribution. Adversarial learning is recently introduced into domain adaptation with promising performance~\cite{DANN,ADDA,cdan,xia2020structure}. By executing generated features to confuse the discriminator, meanwhile forming the discriminator to distinguish from source to target, domain adversarial training aligns features from separate domains through a min-max optimization and delivers domain-invariant representations.

Existing deep domain adaptation methods mainly focus on feature-level alignment. Unfortunately, such ill-judged consideration is not enough to guarantee the success of a beneficial adaptation. As another component of domain shift, the disparity in distributions of labels between source and target domain, \textit{i.e.}, the number of samples in each category differs from source to target, is named as label distribution drift in corresponding to the shift in the distribution of samples across domains. As presented in~\cref{fig:motivation1}, label distribution drift is pernicious according to two aspects. First, along with the adapted processing, features belonging to a large-scale category in target domain are inevitably approaching features in mismatched categories in source domain due to the imbalanced adaptation toward label distribution. As a result, the alignment corrupts feature representations of those misaligned samples. Second, the decision boundary of the classifier is only trained on labeled source samples, and is not applicable to target domain when label distributions differ significantly. These two inside reasons make the adaptation power down under the label distribution drift scenario. The foregoing perspective is supported by practical evidence. \cref{fig:motivation2} shows the performance of the classical domain adversarial method DANN~\cite{DANN} on varying degrees of label distribution drift on \textit{Office-31} dataset~\cite{office31}. Solid and dashed lines represent slight and huge label distribution drift, while all experiments with the same amount of training data. Two observations are quite clear: (1) Compared to the training on the original dataset (black line), the solid red and blue lines deliver similar results, indicating that even if the size of classes in the same domain is imbalanced, high performance can still be achieved by DANN under the scenario of accordant source and target label distribution. Although training sets in solid lines contain different samples, dropped samples do not bring negative effects on the performance; (2) The disparity between solid and dashed lines indicates that when there is a significant label distribution drift between two domains, the performance drops dramatically. The adaptation performance of DANN becomes much worse with a larger divergence between source and target label distribution. More challenging, different from sample distribution shift between domains, label distribution cannot be aligned directly by existing methods because of the unknown target label distribution, and it becomes more difficult to address this problem when a considerable label distribution drift exists.

In this paper, we consider label distribution drift in visual unsupervised domain adaptation, and present a solution whereby managing data shift together with label distribution drift in a unified framework. As mentioned above, domain adaptation with only feature space alignment is only the partial picture. Therefore, we attempt to align two domains on the premise that corresponding label distributions are roughly matched, and continually alleviate data shift and label distribution drift simultaneously during training. To this end, we propose the \textbf{L}abel distribution \textbf{M}atching \textbf{D}omain \textbf{A}dversarial \textbf{N}etwork (LMDAN). We propose a novel weighting strategy for source sample re-weighting, which explores training samples that benefit advantageous adaptation while mitigating negative influences from aligning irrelevant categories across domains. The proposed re-weighting function is capable of contributing to both the adversarial feature alignment and classified boundary learning, hence alleviating the two-fold negative impacts brought by considerable label distribution drift simultaneously.

\section{Related Work}

Feature alignment aims to reduce the domain divergence in feature space. Traditional methods construct projections for two domains, mapping two feature distributions into the manifold space or subspace to address the domain shift problem~\cite{GFK,subspace,dinglowrank,dinglowrank2}. Recently, Long~\textit{et al.}~\cite{dan,JAN} use deep models to reduce the discrepancy between feature spaces in multiple layer levels. Further, by the success of Generative Adversarial Network~\cite{GAN}, adversarial learning in deep models for unsupervised domain adaptation delivers favorable performance~\cite{cdan,bsp,sysnets}. Ganin~\textit{et al.}~\cite{DANN} is the first to employ an adversarial learning-based domain adaptation model and pave to many following works. Tzeng~\textit{et al.}~\cite{ADDA} uses two separate encoders for adaptation while decomposing the transfer process in an end-to-end fashion. Besides, the incorporation of conditional distribution~\cite{cdan,conditional} in adaptation is also a promising way to reach domain-invariant representations. Optimal transport for domain adaptation~\cite{optimaltransportda,mappingestimation,deepjdot,normalizedwass} is another interesting line of research, where source samples are mapped into target domain with minimal cost transportation. Apart from domain adaptation, feature alignment between different groups of samples also has been used in techniques addressing machine learning fairness~\cite{chhabra2022robust,song2021deep,li2020deepfair,li2020dyadic,li2022achieving}. Although great efforts have been made to seek better feature alignment, only handling the feature divergence is not sufficient to guarantee a good adaptation without negative transfer.

Besides the aforementioned challenge and the spring-up of solutions, label distribution drift is another inherent barrier in domain adaptation problems but with less exploration compared to the divergence in feature space. The barrier derives from the divergence between known source label and unknown target label distribution. Previously, some works do indeed formulate label distribution drift problems or so-called target shift~\cite{pmlr-v28-zhang13d}, however, their emphasis is not on deep visual domain adaptation. Liang~\textit{et al.}~\cite{negativetransfer14} focus on the negative transfer and imbalanced distributions in multi-source transfer learning, while Ming~\textit{et al.}~\cite{ming2015unsupervised} exploit label and structural information within and across domains based on the maximum mean discrepancy.

Holding a different emphasis from existing studies, we tackle the deep visual domain adaptation problem under considerable label distribution drift situations and conduct a comprehensive cognition on label distribution drift from both experimental and methodological perspectives. Based on these, we introduce a novel label-matching strategy by continually seeking samples that benefit positive adaptation and simultaneously prevent the negative transfer.

\section{Label Distribution Drift}
\label{sec:motivation}
\vspace{-2mm}

\paragraph{Notation} We start from the basic notations. Consider taking $n_s$ labeled samples $\left\{{(x_i^s,y_i^s)}\right\}^{n_s}_{i=1}$ $\in$ $(\mathcal{X}\times\mathcal{Y})^{n_s}$ under domain distribution $\euscr{D}^s$, where $\mathcal{X}$ and $\mathcal{Y}$ denotes the corresponding feature and label space, and $n_t$ unlabeled samples with the same label set as source samples $\left\{{x_j^t}\right\}^{n_t}_{j=1}$$\in$$\mathcal{X}^{n_t}$ taken from target domain distribution $\euscr{D}^t$. The goal of unsupervised domain adaptation is to utilize labeled source data for the predictions on unlabeled target samples. Suppose an encoder $\mathcal{F}$ that is designed for projecting samples drawn i.i.d. from the input space $\euscr{D}^s$ and $\euscr{D}^t$ to a shared feature space.


\begin{figure}[t]
\centering
\includegraphics[width=0.6\columnwidth]{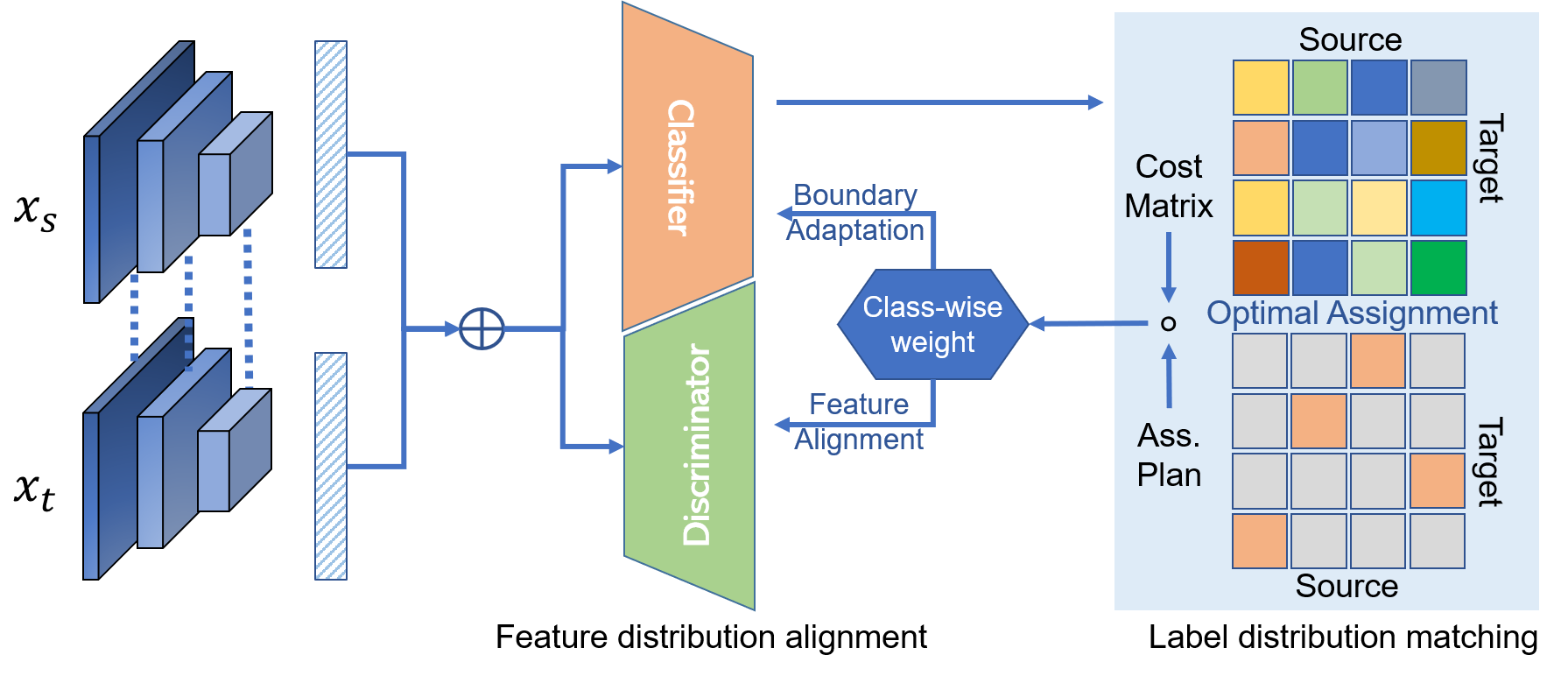}
\setlength{\belowcaptionskip}{-0.5cm}
\caption{Label distribution Matching Domain Adversarial Network.}\label{fig:pipeline}\vspace{-2mm}
\end{figure}

\paragraph{Label Distribution Drift} Tremendous efforts have been made to explore solutions for unsupervised domain adaptation. Unfortunately, most existing studies only focus on feature space divergence by minimizing $Div(\mathbb{E}_{x\sim\euscr{D}^s}[\mathcal{F}(x)], \mathbb{E}_{x\sim\euscr{D}^t}[\mathcal{F}(x)])$, while ignoring the negative effects brought by label distribution drift. Here, $Div(,)$ can be understood as the distribution divergence in terms of any distribution divergence measurement, \textit{e.g.}, KL divergence, or Wasserstein distance.

From a theoretical perspective, a generalization bound for domain adaptation problem towards the expected error on target samples~\cite{ben2010} is given as follows:
\begin{equation}
\begin{split}
    {\epsilon^t}(h)\,\leq{\,}{\epsilon^s}(h)\,+\, {\frac{1}{2}}d_{\mathcal{H}\Delta\mathcal{H}}(\euscr{D}^s,\euscr{D}^t)\,+\, {\lambda^*}\,+\, \Omega,
\end{split}
\end{equation}where ${\epsilon^t}(h)$ and ${\epsilon^s}(h)$ are expected errors on target and source domain, respectively; $\mathcal{H}$ is hypothesis space, $\lambda^*:={\epsilon^s}(h^*)+{\epsilon^t}(h^*)$ is the optimal joint risk among source and target samples, and $\Omega$ is a constant related to the numbers of samples, dimensions, confidence level, and VC-dimension of $\mathcal{H}$. With the assumption that source and target label distribution are close enough, methods with only feature alignment could achieve small target error $\epsilon^t(h)$ by reducing domain distance term $\frac{1}{2}d_{\mathcal{H}\Delta\mathcal{H}}(\euscr{D}^s,\euscr{D}^t)$. However, as pointed out by Zhao~\textit{et al.}~\cite{invariant}, when the above assumption does not hold, the huge label distribution gap between two domains leads the joint error term $\lambda^*$ increase oppositely during the optimization towards the domain distance term, and might counteract with the reduction in domain distance term, which increases the upper bound.

\cref{fig:motivation2} further supports the illustrated statement with practical evidence. Based on the above theoretical analysis and empirical illustration, merely realizing the alignment of feature distribution is still far away from the success of the adaptation. The above exploration motivates us to provide a unified problem formulation of unsupervised domain adaptation on both data distribution shift and label distribution drift.

\paragraph{Problem Formulation} Two challenges for unsupervised domain adaptation problems brought by domain divergence are as follows.

\textbf{Data Distribution Shift}. Usually, samples and extracted feature representations from source to target domain are different, \textit{i.e.}, $\mathcal{P}(x^s)\neq\mathcal{P}(x^t)$, and prohibits models from learning a classifier with labeled source samples that can be directly applied for target sample predictions. For this reason, feature alignment can be achieved by minimizing the divergence:
\begin{equation}
Div(\mathbb{E}_{x^s\sim\euscr{D}_s}[\mathcal{F}(x^s)], \mathbb{E}_{x^t\sim\euscr{D}_t}[\mathcal{F}(x^t)])\ ,
\end{equation}or by minimizing the distribution divergence that is conditional on the categorical belonging of samples to narrow the data distribution shift for domain adaptation:
\begin{equation}
Div(\mathbb{E}_{(x^s,y^s)\sim\euscr{D}^s}[\mathcal{F}(x^s)|y^s],\mathbb{E}_{(x^t,y^t)\sim\euscr{D}^t}[\mathcal{F}(x^t)|y^t]),\ \forall{y^s=y^t}.
\end{equation}

\textbf{Label Distribution Drift}. Beyond the inconsistency in feature space, domain divergence also occurs in label space, where $\mathcal{P}(y^s)\neq\mathcal{P}(y^t)$. It is more challenging to handle label distribution drift than data shift due to $\mathcal{P}(y^t)$ being an agnostic distribution in unsupervised domain adaptation.

When considerable label distribution divergence exists, since we always equally sample from two domains and feed the sampling into adversarial alignment, the excessive training towards feature distribution alignment will mislead to the unaligned minimization:
\begin{equation}
Div(\mathbb{E}_{(x^s,y^s)\sim\euscr{D}^s}[\mathcal{F}(x^s)|y^s], \mathbb{E}_{(x^t,y^t)\sim\euscr{D}^t}[\mathcal{F}(x^t)|y^t]),\ \forall{y^s\neq{y^t}},
\end{equation}which aligns target feature representations to irrelevant categories in source domain during training. These can induce corruption in the level of categorical feature representations and rise increasing predicted error when predicting $y^t$ in target domain, and further turn out negative effects on the adaptation.

Most deep visual unsupervised domain adaptation methods consider domain divergence merely from data shift, and take no drift in label space for granted. We need to admit that addressing the partial picture is enough for experimental datasets, since current datasets do not include a considerable label distribution drift. We lament that such success is not only far away from real scenarios but also suffers from degraded performance due to label distribution drift (See~\cref{fig:motivation2}). Consider adverse effects can be easily accessed when the inconsistent label distribution and the alignment between irrelevant classes across domains exist, not all samples can be fully used during the training process for the reason that positive adaptation only comes from the part of a correctly matched pair of source-target samples. Consequently, we try to exploit and emphasize the part of correctly matched samples in two domains, while mitigating the alignment on class-mismatched samples. Thus we further increase the ratio of positive adaptation and avoid the negative transfer brought by the alignment between irrelevant categories across source and target domains. This can be viewed as an unsupervised sample selection in that we are continually seeking samples that benefit from adaptation and avoid negative transfer concurrently.

\section{Label Distribution Matching}

\paragraph{Overview} \cref{fig:pipeline} shows the framework of the proposed LMDAN. It minimizes the domain divergence embedded in feature space on the premise of close source-target label distribution. Specifically, to align the source and target domain under label distribution drift, LMDAN contains two interactive parts: adversarial training for domain-invariant features generation, and the class-wise re-weighting strategy through the optimal assignment for source sample selection. In adversarial feature alignment, the encoder $\mathcal{F}$ tries to extract feature $\mathcal{F}(x^s)$ and $\mathcal{F}(x^t)$ from two domains and confuse the discriminator $\mathcal{D}$, while $\mathcal{D}$ tries to distinguish $\mathcal{F}(x^s)$ and $\mathcal{F}(x^t)$ from each other. Finally, $\mathcal{F}$ is trained to map data distribution from two domains close enough. In the source samples weighting part, by adding class-wise weights on both adversarial training and supervision on the classifier, we manipulate feature alignment in adversarial training and decision boundary of the classifier to tackle the label drift scenario simultaneously. The dual weighting strategy makes the network adapt to target domain by two sides: (1) The weighting for the min-max game emphasizes features in the same category to get closer across domains, at the same time mitigate the misalignment, and (2) The weighting for classifier makes decision boundary adapt to the target label distribution. In the following, we emphatically illustrate the source sample weighting strategy, and then provide details for model training.


\subsection{Label Distribution Matching}

Label distribution matching is one of the crucial components in the LMDAN framework. It disposes of label distribution drift towards source-target sample matching. Here, we expect to exploit samples in the parts of classes matched across source and target domain by the optimal assignment, then enlarge matched classes and shrink the size of less relevant classes in source domain. As a result, samples in source domain engaging in the adversarial feature alignment are able to approach to target domain in terms of label distribution, and further contribute to increasing positive and mitigating negative transfer.

To achieve this, we employ the classified probability $g$ with $||g||_1$$=$$1$ of every sample to measure the degree of matching. Based on the measurement of distance and optimal matching, mismatched pairs result in a larger distance, while matched pairs perform inversely. Consider a cost function $c:\mathcal{C}\times\mathcal{C}\to\mathbb{R}^{+}$ and $g_i^s$ and $g_j^t$ the classified probabilities obtained by the classifier $\mathcal{G}$ for source sample $x_i^s$ and target sample $x_j^t$, respectively, and the output space $\mathcal{C}:g\in\mathcal{C}$. Based on optimal assignment~\cite{optimaltransport}, we seek for a joint probability distribution $\gamma$ according to $g^s$ and $g^t$:
\begin{equation}
\gamma^*\,=\,\mathop{\arg\min}_{\gamma\in\prod(\mathcal{C}\times\mathcal{C})}\int_{\mathcal{C}\times\mathcal{C}}c(g^s,g^t)d\gamma(g^s,g^t)\ .
\end{equation}
This indicates the optimal assignment based on classified probabilities from source to target with the minimum cost.

As for the discrete version for implementation, we employ Euclidean distance to build the cost matrix $M=\{m_{ij}\}\in\mathbb{R}^{{n_s}\times{n_t}}$ between source and target domain,
\begin{equation}
    m_{ij} = c(g_i^s,g_j^t) = ||g_i^s-g_j^t||_2\ ,
\end{equation}
and other distance functions can be used as well. Based on the cost matrix $M$, the optimal assignment is written as:
\begin{equation}\label{eq:trans}
\vspace{-1.0mm}
    \gamma^*=\mathop{\arg\min}_{\gamma\in\mathbb{R}^{n_s\times{n_t}}}\left\langle\gamma,M\right\rangle_F, \ \textup{s.t.}:\gamma\textbf{1}_{n_t}=\sum^{n_s}_i{g^s_i},\ \gamma\textbf{1}_{n_s}=\sum^{n_t}_j{g^t_j}\ ,
\end{equation}
where $\left\langle\cdot,\cdot\right\rangle_F$ indicates Frobenius inner product, and $\textbf{1}_{n}$ is an all-one $n$-dimension vector.


We then incorporate the distance within classified probabilities into the optimal assignment plan and make the conjunct term guide class-wise weights for each class. We obtain the weight guiding matrix $\textbf{T}=\{t_{ij}\}\in\mathbb{R}^{n_s\times{n_t}}$ by
\begin{equation}\label{eq:product}
\textbf{T} = \gamma^*\,\circ\,\textbf{M}\ ,
\end{equation}
\noindent where $\circ$ denotes the Hadamard product. By matching classification probabilities with the minimal cost, the weight guiding matrix provides guidance for misaligned samples. Moreover, following the above step, we compute the class-wise weight $w_k$ for the class with index $k$ in source domain by:
\begin{equation}\label{eq:weight}
w_k = \Big(({\sum_{i=1}^{n_s}\mathbbm{1}_{y_i^s=k}})^\alpha\cdot{\sum_{i=1}^{n_s}t_{ij}\mathbbm{1}_{y_i^s=k}}\Big)^{-1},
\end{equation}
where $\mathbbm{1}$ is the indicator function. Note that $w_k$ consists of two parts, where the first term manages the imbalanced class size within source domain itself, and the second awards or punishes the matched or mismatched pairs between source and target accordingly. $\alpha$ is the parameter to control the influence of source class imbalanced scale.

Using weights in terms of categories according to the optimal matching toward classified probabilities, we are able to distinguish classes that are misaligned and less relevant to positive transfer from well-aligned ones during training. By re-weighting samples in source domain by class-wise weights, the sizes or corresponding categories are enlarged or shrunk accordingly, then further push the source label distribution to the unknown target one dynamically.


\subsection{Objective and Solution}

Finally, we provide the objective functions and the corresponding optimizing solution for LMDAN. We first calculate class-wise weights on each mini-batch sample and then optimize toward the min-max game in adversarial learning integrated with subsequent classification by a dual weighting strategy. Loss functions for LMDAN can be written as:
\begin{equation}\label{eq:loss}
\min_{\mathcal{F},\mathcal{G}}\  \mathcal{L}_1 (\mathcal{F},\mathcal{G},\mathcal{D})+ \lambda \mathcal{L}_2(\mathcal{F},\mathcal{D})\ \ \textup{and}\  \max_{\mathcal{D}}\mathcal{L}_2(\mathcal{F},\mathcal{D}), \ \textup{with} \\
\end{equation}
\begin{equation}\nonumber 
\begin{split}
    &\mathcal{L}_1= \underset{(x^s,y^s)\sim{\euscr{D}^s}}{\mathbb{E}} w_i\mathcal{L}\,(\mathcal{G}\,(\mathcal{F}(x^s)),\,y^s),\\
    &\mathcal{L}_2= \underset{x^s\sim{\euscr{D}^s}}{\mathbb{E}} w_i\log\left[\mathcal{D}\,(\mathcal{F}\,(x^s))\right]+ \underset{x^t\sim{\euscr{D}^t}}{\mathbb{E}}\log\left[1-\mathcal{D}\,(\mathcal{F}\,(x^t))\right],
\end{split}
\end{equation}where $w_i$ is the corresponding weight of the class where $x^s\sim\euscr{D}^s$ belongs to, and $\lambda$ is the trade-off hyperparameter for classification loss and adversarial loss. In our objective functions, the weighting strategy is conducted in two places. The weighted classifier $\mathcal{G}$ captures label distribution drift for better decision boundary adaptation on target domain, and the weighted discriminator $\mathcal{D}$ and encoder $\mathcal{F}$ further adjust feature alignment to fit label distribution drift as well.

In our implementation, we utilize cross-entropy loss as the loss function for $\mathcal{L}$, and set the trade-off parameter $\lambda$ default to 1 for all experiments. Since the complexity of the optimal assignment is not scalable to the whole dataset, the mini-batch label matching is applied as well. Two benefits are clear. Mini-batch training makes the complexity of the optimal matching affordable in big data adaptation. Besides, equivalent numbers of data points from source and target domain can be sampled, rendering the matching and feature alignment balanced. We use pre-trained ResNet-50~\cite{resnet} as the feature extractor. Following by~\cite{DANN}, we set the initial learning rate $lr=0.01$ for SGD optimizer, then gradually adjust the learning rate for the classifier by ${lr_{c} = lr(1+10p)^{-0.75}}$, where $p$ is the training process changed from $0$ to $1$ linearly. The learning rate for discriminator is $lr_{d}=\frac{1-\exp(-10p)}{1+\exp(-10p)}lr$.

\vspace{-3mm}
\section{Experimental Analysis}
\label{sec:exp}

\begin{table*}[t]
  \centering
  \caption{Results for unsupervised domain adaptation with label distribution drift on \textit{Office-31}[0.75;0.75] dataset.}
  \label{tab:office}
  \resizebox{1.\columnwidth}{!}{
    \begin{tabular}{cccccccc}
    \toprule
    Method & \textbf{A} $\rightarrow$ \textbf{W} & \textbf{A} $\rightarrow$ \textbf{D} & \textbf{W} $\rightarrow$ \textbf{A} & \textbf{W} $\rightarrow$ \textbf{D} & \textbf{D} $\rightarrow$ \textbf{A} & \textbf{D} $\rightarrow$ \textbf{W} & Average \\
    \midrule
    ResNet50~\cite{resnet} & 66.1 $\pm$ 4.3 & 65.8 $\pm$ 1.5 & 53.3 $\pm$ 3.1 & \textbf{87.8} $\pm$ \textbf{2.9} & 53.0 $\pm$ 3.7 & 79.4 $\pm$ 2.6 & 67.6 $\pm$ 1.5 \\
    DANN~\cite{DANN} & 50.7 $\pm$ 2.6 & 54.0 $\pm$ 2.7 & 35.4 $\pm$ 3.4 & 62.6 $\pm$ 4.2 & 34.6 $\pm$ 3.8 & 56.3 $\pm$ 2.9 & 49.0 $\pm$ 0.8 \\
    JAN~\cite{JAN} & 51.2 $\pm$ 3.2 & 49.5 $\pm$ 2.4 & 46.1 $\pm$ 3.9 & 72.9 $\pm$ 4.1 & 40.9 $\pm$ 5.1 & 71.8 $\pm$ 2.6 & 55.4 $\pm$ 1.6 \\
    WMMD~\cite{wmmd} & 39.1 $\pm$ 5.2 & 43.3 $\pm$ 4.1 & 38.4 $\pm$ 2.7 & 67.8 $\pm$ 4.8 & 34.1 $\pm$ 3.2 & 68.1 $\pm$ 7.1 & 48.5 $\pm$ 3.4 \\
    CDAN~\cite{cdan} & 65.7 $\pm$ 3.2 & 62.8 $\pm$ 4.8 & 52.5 $\pm$ 2.7 & 78.1 $\pm$ 4.7 & 39.8 $\pm$ 4.5 & 73.5 $\pm$ 4.4 & 62.1 $\pm$ 1.7 \\
    RAAN~\cite{RAAN} & 59.4 $\pm$ 3.8 & 65.7 $\pm$ 2.9 & 48.5 $\pm$ 5.0 & 76.4 $\pm$ 3.5 & 45.8 $\pm$ 6.9 & 77.4 $\pm$ 3.6 & 62.2 $\pm$ 3.2 \\
    SymNets~\cite{sysnets} & 57.1 $\pm$ 4.0 & 54.6 $\pm$ 2.7 & 41.9 $\pm$ 6.3 & 67.0 $\pm$ 5.1 & 32.4 $\pm$ 4.8 & 57.2 $\pm$ 6.7 & 51.7 $\pm$ 2.7 \\
    BSP~\cite{bsp} & 61.5 $\pm$ 2.1 & 58.9 $\pm$ 2.6 & 47.5 $\pm$ 3.2 & 85.0 $\pm$ 3.6 & 40.4 $\pm$ 2.9 & 84.1 $\pm$ 3.0 & 62.9 $\pm$ 2.2 \\
    \midrule
    LMDAN & \textbf{73.1} $\pm$ \textbf{1.7} & \textbf{71.0} $\pm$ \textbf{2.5} & \textbf{56.5} $\pm$ \textbf{2.4} & 84.4 $\pm$ 2.6 & \textbf{57.8} $\pm$ \textbf{4.9} & \textbf{88.8} $\pm$ \textbf{3.5} & \textbf{71.9 $\pm$ 2.1} \\
    \bottomrule
    \end{tabular}
    }\vspace{-6mm}
\end{table*}


\begin{table*}[t]
  \centering
  \caption{Results for unsupervised domain adaptation with label distribution drift on \textit{ImageCLEF-DA}[0.75;0.75] dataset.}
  \label{tab:imageclef}
    \resizebox{1.\columnwidth}{!}{
    \begin{tabular}{cccccccc}
    \toprule
    Method & \textbf{C} $\rightarrow$ \textbf{I} & \textbf{C} $\rightarrow$ \textbf{P} & \textbf{I} $\rightarrow$ \textbf{C} & \textbf{I} $\rightarrow$ \textbf{P} & \textbf{P} $\rightarrow$ \textbf{C} & \textbf{P} $\rightarrow$ \textbf{I} & Average \\
    \midrule
    ResNet50~\cite{resnet} & 76.9 $\pm$ 3.2 & 63.8 $\pm$ 1.7 & 87.1 $\pm$ 1.8 & 71.3 $\pm$ 0.7 & 81.7 $\pm$ 3.7 & 73.4 $\pm$ 4.2 & 75.7 $\pm$ 2.1 \\
    DANN~\cite{DANN} & 47.4 $\pm$ 2.8 & 40.8 $\pm$ 2.7 & 55.0 $\pm$ 1.2 & 50.4 $\pm$ 2.3 & 55.0 $\pm$ 3.1 & 51.2 $\pm$ 3.6 & 50.0 $\pm$ 1.4 \\
    JAN~\cite{JAN} & 34.2 $\pm$ 2.8 & 27.9 $\pm$ 1.0 & 38.8 $\pm$ 3.8 & 49.0 $\pm$ 3.4 & 36.7 $\pm$ 4.4 & 44.1 $\pm$ 3.0 & 38.5 $\pm$ 0.7 \\
    WMMD~\cite{wmmd} & 42.4 $\pm$ 1.1 & 30.4 $\pm$ 3.5 & 65.2 $\pm$ 3.9 & 70.8 $\pm$ 3.0 & 47.2 $\pm$ 3.5 & 56.4 $\pm$ 1.9 & 52.0 $\pm$ 2.9 \\
    CDAN~\cite{cdan} & 58.1 $\pm$ 3.8 & 52.2 $\pm$ 3.2 & 76.3 $\pm$ 4.0 & 62.7 $\pm$ 1.8 & 66.2 $\pm$ 9.5 & 59.2 $\pm$ 1.2 & 63.1 $\pm$ 2.2 \\
    RAAN~\cite{RAAN} & 62.9 $\pm$ 1.3 & 54.6 $\pm$ 3.3 & 78.3 $\pm$ 1.7 & 63.6 $\pm$ 3.6 & 71.0 $\pm$ 6.6 & 65.4 $\pm$ 2.4 & 66.0 $\pm$ 2.3 \\
    SymNets~\cite{sysnets} & 59.2 $\pm$ 5.0 & 53.8 $\pm$ 3.2 & 70.5 $\pm$ 3.9 & 57.2 $\pm$ 3.8 & 63.4 $\pm$ 7.6 & 54.3 $\pm$ 1.5 & 59.7 $\pm$ 1.4 \\
    BSP~\cite{bsp} & 52.6 $\pm$ 1.7 & 43.4 $\pm$ 2.5 & 70.5 $\pm$ 2.9 & 58.6 $\pm$ 4.6 & 67.0 $\pm$ 4.3 & 62.9 $\pm$ 1.8 & 59.2 $\pm$ 1.8 \\
    \midrule
    LMDAN & \textbf{79.1 $\pm$ 2.8} & \textbf{67.7 $\pm$ 2.7} & \textbf{89.8 $\pm$ 2.3} & \textbf{71.6 $\pm$ 2.8} & \textbf{88.1 $\pm$ 2.4} & \textbf{80.5 $\pm$ 1.0} & \textbf{79.5 $\pm$ 0.8} \\
    \bottomrule
  \end{tabular}
  }\vspace{-6mm}
\end{table*}


Due to the space limitation, we defer some experimental settings and most of the experimental results (more benchmarking results, ablation studies, and visualization) to supplementary materials.

\cref{tab:office} reports quantitative results for unsupervised domain adaptation on \textit{Office-31}[0.75;0.75] dataset. The performance of all competitive methods significantly drops under the huge label distribution divergence and even becomes worse than non-adapted ResNet-50. This indicates that only aligning the feature divergence is not enough for a positive adaptation, for label distribution drift is also a crucial component of domain shift, and has not been paid sufficient attention in domain adaptation area. To dispose of this problem, LMDAN considers source-target sample pairs with different weights, enlarges and shrinks weights for matched and mismatched samples on classified probabilities. By this means, LMDAN outperforms other competitive methods by a large margin. To be noticed, WMMD and RAAN also embed source sample re-weighting strategies into training. However, their weights highly rely on predictions of the target label distribution and make them struggle to handle huge label distribution drift.


\vspace{-4mm}
\section{Conclusion}

We proposed the Label distribution Matching Domain Adversarial Network (LMDAN) framework for unsupervised domain adaptation. We designed the label distribution matching and weighting strategy for source samples re-weighting, and matched the known source label distribution with the agnostic target one. Experimental results demonstrated the superior performance of LMDAN over other state-of-the-art methods.

%
%
%
%

\vspace{-4mm}
\bibliographystyle{splncs04}
\bibliography{egbib}

\begin{thebibliography}{10}
\providecommand{\url}[1]{\texttt{#1}}
\providecommand{\urlprefix}{URL }
\providecommand{\doi}[1]{https://doi.org/#1}

\bibitem{normalizedwass}
Balaji, Y., Chellappa, R., Feizi, S.: Normalized wasserstein for mixture
  distributions with applications in adversarial learning and domain
  adaptation. In: Proceedings of The IEEE International Conference on Computer
  Vision (2019)

\bibitem{ben2010}
Ben-David, S., Blitzer, J., Crammer, K., Kulesza, A., Pereira, F., Vaughan,
  J.W.: A theory of learning from different domains. Machine Learning  (2010)

\bibitem{deepjdot}
Bhushan~Damodaran, B., Kellenberger, B., Flamary, R., Tuia, D., Courty, N.:
  Deepjdot: Deep joint distribution optimal transport for unsupervised domain
  adaptation. In: Proceedings of The European Conference on Computer Vision
  (2018)

\bibitem{RAAN}
Chen, Q., Liu, Y., Wang, Z., Wassell, I., Chetty, K.: Re-weighted adversarial
  adaptation network for unsupervised domain adaptation. In: Proceedings of The
  IEEE Conference on Computer Vision and Pattern Recognition (2018)

\bibitem{bsp}
Chen, X., Wang, S., Long, M., Wang, J.: Transferability vs. discriminability:
  Batch spectral penalization for adversarial domain adaptation. In:
  Proceedings of the 36th International Conference on Machine Learning.
  Proceedings of Machine Learning Research (2019),
  \url{http://proceedings.mlr.press/v97/chen19i.html}

\bibitem{chhabra2022robust}
Chhabra, A., Li, P., Mohapatra, P., Liu, H.: Robust fair clustering: A novel
  fairness attack and defense framework. In: The Eleventh International
  Conference on Learning Representations (2022)

\bibitem{conditional}
Cicek, S., Soatto, S.: Unsupervised domain adaptation via regularized
  conditional alignment. In: Proceedings of The IEEE International Conference
  on Computer Vision (October 2019)

\bibitem{optimaltransportda}
Courty, N., Flamary, R., Tuia, D., Rakotomamonjy, A.: Optimal transport for
  domain adaptation. IEEE Transactions on Pattern Analysis and Machine
  Intelligence  (2017). \doi{10.1109/TPAMI.2016.2615921}

\bibitem{8398464}
{Ding}, Z., {Nasrabadi}, N.M., {Fu}, Y.: Semi-supervised deep domain adaptation
  via coupled neural networks. IEEE Transactions on Image Processing
  \textbf{27}(11),  5214--5224 (2018)

\bibitem{ding2017deep}
Ding, Z., Fu, Y.: Deep domain generalization with structured low-rank
  constraint. IEEE Transactions on Image Processing  \textbf{27}(1),  304--313
  (2017)

\bibitem{dinglowrank2}
Ding, Z., Fu, Y.: Deep transfer low-rank coding for cross-domain learning. IEEE
  Transactions on Neural Networks and Learning Systems  (2019)

\bibitem{Ding_2018_ECCV}
Ding, Z., Li, S., Shao, M., Fu, Y.: Graph adaptive knowledge transfer for
  unsupervised domain adaptation. In: Proceedings of the European Conference on
  Computer Vision (ECCV) (September 2018)

\bibitem{dingzeroshot}
Ding, Z., Liu, H.: Marginalized latent semantic encoder for zero-shot learning.
  In: Proceedings of The IEEE Conference on Computer Vision and Pattern
  Recognition (2019)

\bibitem{dinglowrank}
Ding, Z., Shao, M., Fu, Y.: Deep low-rank coding for transfer learning. In:
  Twenty-Fourth International Joint Conference on Artificial Intelligence
  (2015)

\bibitem{ding2018robust}
Ding, Z., Shao, M., Fu, Y.: Robust multi-view representation: A unified
  perspective from multi-view learning to domain adaption. In: IJCAI. pp.
  5434--5440 (2018)

\bibitem{Ding2019}
Ding, Z., Zhao, H., Fu, Y.: Deep Domain Adaptation, pp. 203--249. Springer
  International Publishing, Cham (2019)

\bibitem{subspace}
Fernando, B., Habrard, A., Sebban, M., Tuytelaars, T.: Unsupervised visual
  domain adaptation using subspace alignment. In: Proceedings of The IEEE
  International Conference on Computer Vision (2013)

\bibitem{DANN}
Ganin, Y., Ustinova, E., Ajakan, H., Germain, P., Larochelle, H., Laviolette,
  F., Marchand, M., Lempitsky, V.: Domain-adversarial training of neural
  networks. Journal of Machine Learning Research  (2016)

\bibitem{negativetransfer14}
Ge, L., Gao, J., Ngo, H., Li, K., Zhang, A.: On handling negative transfer and
  imbalanced distributions in multiple source transfer learning. Statistical
  Analysis and Data Mining: The ASA Data Science Journal  (2014)

\bibitem{GFK}
Gong, B., Shi, Y., Sha, F., Grauman, K.: Geodesic flow kernel for unsupervised
  domain adaptation. In: Proceedings of The IEEE Conference on Computer Vision
  and Pattern Recognition (2012). \doi{10.1109/CVPR.2012.6247911}

\bibitem{GAN}
Goodfellow, I., Pouget-Abadie, J., Mirza, M., Xu, B., Warde-Farley, D., Ozair,
  S., Courville, A., Bengio, Y.: Generative adversarial nets. In: Advances in
  Neural Information Processing Systems (2014)

\bibitem{resnet}
He, K., Zhang, X., Ren, S., Sun, J.: Deep residual learning for image
  recognition. In: Proceedings of The IEEE Conference on Computer Vision and
  Pattern Recognition (2016)

\bibitem{optimaltransport}
Kantorovich, L.V.: On the translocation of masses. Journal of Mathematical
  Sciences  (2006)

\bibitem{zeroshot}
Kodirov, E., Xiang, T., Fu, Z., Gong, S.: Unsupervised domain adaptation for
  zero-shot learning. In: Proceedings of The IEEE International Conference on
  Computer Vision (2015)

\bibitem{li2021selfdoc}
Li, P., Gu, J., Kuen, J., Morariu, V.I., Zhao, H., Jain, R., Manjunatha, V.,
  Liu, H.: Selfdoc: Self-supervised document representation learning. In:
  Proceedings of the IEEE/CVF Conference on Computer Vision and Pattern
  Recognition. pp. 5652--5660 (2021)

\bibitem{li2022achieving}
Li, P., Liu, H.: Achieving fairness at no utility cost via data reweighing with
  influence. In: International Conference on Machine Learning. pp.
  12917--12930. PMLR (2022)

\bibitem{li2022exploiting}
Li, P., Wang, P., Berntorp, K., Liu, H.: Exploiting temporal relations on radar
  perception for autonomous driving. In: Proceedings of the IEEE/CVF Conference
  on Computer Vision and Pattern Recognition. pp. 17071--17080 (2022)

\bibitem{li2020dyadic}
Li, P., Wang, Y., Zhao, H., Hong, P., Liu, H.: On dyadic fairness: Exploring
  and mitigating bias in graph connections. In: International Conference on
  Learning Representations (2020)

\bibitem{li2020deepfair}
Li, P., Zhao, H., Liu, H.: Deep fair clustering for visual learning. In:
  Proceedings of the IEEE/CVF Conference on Computer Vision and Pattern
  Recognition. pp. 9070--9079 (2020)

\bibitem{8362753}
{Li}, S., {Song}, S., {Huang}, G., {Ding}, Z., {Wu}, C.: Domain invariant and
  class discriminative feature learning for visual domain adaptation. IEEE
  Transactions on Image Processing  \textbf{27}(9),  4260--4273 (2018)

\bibitem{li2020deep}
Li, S., Liu, C.H., Lin, Q., Wen, Q., Su, L., Huang, G., Ding, Z.: Deep residual
  correction network for partial domain adaptation. IEEE Transactions on
  Pattern Analysis and Machine Intelligence  (2020)

\bibitem{liumultisource}
Liu, H., Shao, M., Fu, Y.: Structure-preserved multi-source domain adaptation.
  In: Proceedings of The IEEE 16th International Conference on Data Mining
  (2016)

\bibitem{dan}
Long, M., Cao, Y., Wang, J., Jordan, M.I.: Learning transferable features with
  deep adaptation networks. In: Proceedings of the 32nd International
  Conference on Machine Learning (2015),
  \url{http://dl.acm.org/citation.cfm?id=3045118.3045130}

\bibitem{cdan}
Long, M., Cao, Z., Wang, J., Jordan, M.I.: Conditional adversarial domain
  adaptation. In: Advances in Neural Information Processing Systems (2018)

\bibitem{JAN}
Long, M., Zhu, H., Wang, J., Jordan, M.I.: Deep transfer learning with joint
  adaptation networks. In: Proceedings of The 34th International Conference on
  Machine Learning (2017)

\bibitem{tsne}
Maaten, L.v.d., Hinton, G.: Visualizing data using t-sne. Journal of Machine
  Learning Research  (2008)

\bibitem{ming2015unsupervised}
Ming Harry~Hsu, T., Yu~Chen, W., Hou, C.A., Hubert~Tsai, Y.H., Yeh, Y.R.,
  Frank~Wang, Y.C.: Unsupervised domain adaptation with imbalanced cross-domain
  data. In: Proceedings of The IEEE International Conference on Computer Vision
  (2015)

\bibitem{peng2017visda}
Peng, X., Usman, B., Kaushik, N., Hoffman, J., Wang, D., Saenko, K.: Visda: The
  visual domain adaptation challenge. arXiv preprint arXiv:1710.06924  (2017)

\bibitem{mappingestimation}
Perrot, M., Courty, N., Flamary, R., Habrard, A.: Mapping estimation for
  discrete optimal transport. In: Advances in Neural Information Processing
  Systems (2016)

\bibitem{office31}
Saenko, K., Kulis, B., Fritz, M., Darrell, T.: Adapting visual category models
  to new domains. In: Proceedings of The European Conference on Computer Vision
  (2010)

\bibitem{song2021deep}
Song, H., Li, P., Liu, H.: Deep clustering based fair outlier detection. In:
  Proceedings of the 27th ACM SIGKDD Conference on Knowledge Discovery \& Data
  Mining. pp. 1481--1489 (2021)

\bibitem{ADDA}
Tzeng, E., Hoffman, J., Saenko, K., Darrell, T.: Adversarial discriminative
  domain adaptation. In: Proceedings of The IEEE International Conference on
  Computer Vision (2017)

\bibitem{xia2020structure}
Xia, H., Ding, Z.: Structure preserving generative cross-domain learning. In:
  Proceedings of the IEEE/CVF Conference on Computer Vision and Pattern
  Recognition. pp. 4364--4373 (2020)

\bibitem{wmmd}
Yan, H., Ding, Y., Li, P., Wang, Q., Xu, Y., Zuo, W.: Mind the class weight
  bias: Weighted maximum mean discrepancy for unsupervised domain adaptation.
  In: Proceedings of The IEEE Conference on Computer Vision and Pattern
  Recognition (2017)

\bibitem{pmlr-v28-zhang13d}
Zhang, K., Schölkopf, B., Muandet, K., Wang, Z.: Domain adaptation under
  target and conditional shift. In: Dasgupta, S., McAllester, D. (eds.)
  Proceedings of the 30th International Conference on Machine Learning (2013)

\bibitem{sysnets}
Zhang, Y., Tang, H., Jia, K., Tan, M.: Domain-symmetric networks for
  adversarial domain adaptation. In: Proceedings of The IEEE Conference on
  Computer Vision and Pattern Recognition (2019)

\bibitem{invariant}
Zhao, H., Combes, R.T.D., Zhang, K., Gordon, G.: On learning invariant
  representations for domain adaptation. In: Proceedings of The 36th
  International Conference on Machine Learning (2019)

\end{thebibliography}

\clearpage
\appendix

\section{Dataset and Modification}

Three widely used real-world datasets are employed to evaluate the performance of LMDAN and other competitive methods. (1) \textit{Office-31}~\cite{office31} contains 4,652 images in total within 31 categories. The dataset contains three domains: \textit{Amazon} (\textbf{A}), \textit{Webcam} (\textbf{W}), and \textit{DSLR} (\textbf{D}), where images are broadly taken from the internet to real scenarios. (2) \textit{Visda-2017}~\cite{peng2017visda} is a challenging domain adaptation dataset, aiming to transfer knowledge from synthetic images (\textbf{S}) to real images (\textbf{R}). It contains around 152,000 synthetic images of 3D models for source domain, and 72,000 real images for target domain. Both two domains contain 12 categories. (3) \textit{ImageCLEF-DA} contains 600 images per domain taken from three object recognition datasets, \textit{Caltech-256} (\textbf{C}), \textit{ImageNet ILSVRC 2012} (\textbf{I}), and \textit{Pascal VOC 2012} (\textbf{P}). In the following experiments, ``\textbf{A}$\rightarrow$\textbf{W}" denotes domain adaptation from source domain \textit{Amazon} to target domain \textit{Webcam}.

\begin{figure}[t]
\centering
\includegraphics[width=0.95\columnwidth]{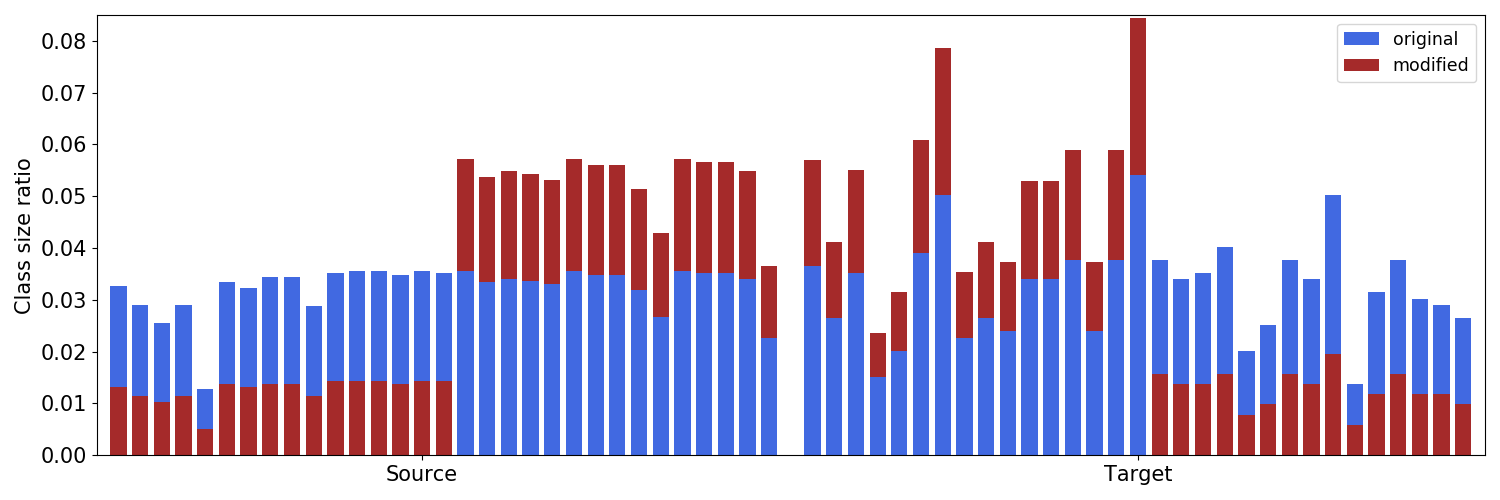}
\setlength{\belowcaptionskip}{-0.5cm}
\caption{\small{Visualization of label distribution on 31 categories on original \textit{Office-31} and modified \textit{Office-31}[0.75;0.75] dataset. Label distribution drift in current datasets is inconspicuous and not enough for us to reveal the negative influence induced by it, so we simulate the scenario by randomly dropping samples within certain categories.}}\label{fig:dataset}
\end{figure}

Due to the label distribution drift in current datasets being inconspicuous, we simulate the huge label distribution drift by randomly dropping out 75\% samples in the first half of classes within source domain, and 75\% samples in the latter half of classes in target domain, and noting the modified dataset as ``NAME[0.75;0.75]'', where NAME is the name of original datasets. \cref{fig:dataset} shows the label distribution on original \textit{Office-31} and modified \textit{Office-31}[0.75;0.75], respectively. The sample-dropping process with randomness is repeated five times, and we conduct experiments on all created datasets while reporting the average performance and its fluctuation to alleviate the sample selection bias.

\section{Competitive Methods}

We compare seven representative or recent deep visual unsupervised domain adaptation methods as well as ResNet50 trained only on source domain without adaptation. JAN~\cite{JAN} and WMMD~\cite{wmmd} are deep transfer models based on the maximum mean discrepancy. They learn the adaptation by aligning joint distributions and minimizing the feature divergence within multiple domain-specific layers. DANN~\cite{DANN}, CDAN~\cite{cdan}, SymNets~\cite{sysnets}, and BSP~\cite{bsp} are based on adversarial training that attempt to make two feature spaces confuse the discriminator. Beyond the somehow arbitrary feature alignment, CDAN~\cite{cdan} and SymNets~\cite{sysnets} take the conditional feature distribution into consideration, which makes the alignment conditioned on the predicted category, and enhances the adapted performance under normal domain adaptation setting. Unlikely with adversarial domain adaptation, RAAN~\cite{RAAN} reduces the divergence in feature distribution by minimizing Earth-Mover distance. For all the competitive methods, we use ResNet50 as the feature extractor for apples-to-apples comparisons. We re-implement WMMD and RAAN, and conduct experiments for the rest of the methods with their open-source codes.

\begin{table}
  \centering
  \caption{Results for unsupervised domain adaptation on \textbf{S}$\rightarrow$\textbf{R} of \textit{VisDA-2017} dataset in original and [0.75;0.75].}\label{tab:visda}
  \fontsize{9.5}{10.0}\selectfont
    \begin{tabular}{ccc}
    \toprule
    Method & Original & [0.75;0.75] \\
    \midrule
    ResNet-50~\cite{resnet} & 44.4 & 41.0 $\pm$ 0.7 \\
    DANN~\cite{DANN} & 63.5 & 33.9 $\pm$ 1.3 \\
    JAN~\cite{JAN} & 61.6 & 27.7 $\pm$ 2.0 \\
    WMMD~\cite{wmmd} & 45.8 & 28.4 $\pm$ 2.2 \\
    CDAN~\cite{cdan} & \textbf{66.8} & 38.5 $\pm$ 0.9 \\
    RAAN~\cite{RAAN} & 59.0 & 50.9 $\pm$ 1.9 \\
    SymNets~\cite{sysnets} & 51.6 & 22.6 $\pm$ 0.7 \\
    BSP~\cite{bsp} & 64.7 & 27.9 $\pm$ 3.2 \\
    \midrule
    LMDAN & 64.9 & \textbf{59.3 $\pm$ 1.1} \\
    \bottomrule
  \end{tabular}
\end{table}

\begin{figure*}[t]
\centering 
\begin{subfigure}{0.32\columnwidth}
    \includegraphics[width=1.0\columnwidth]{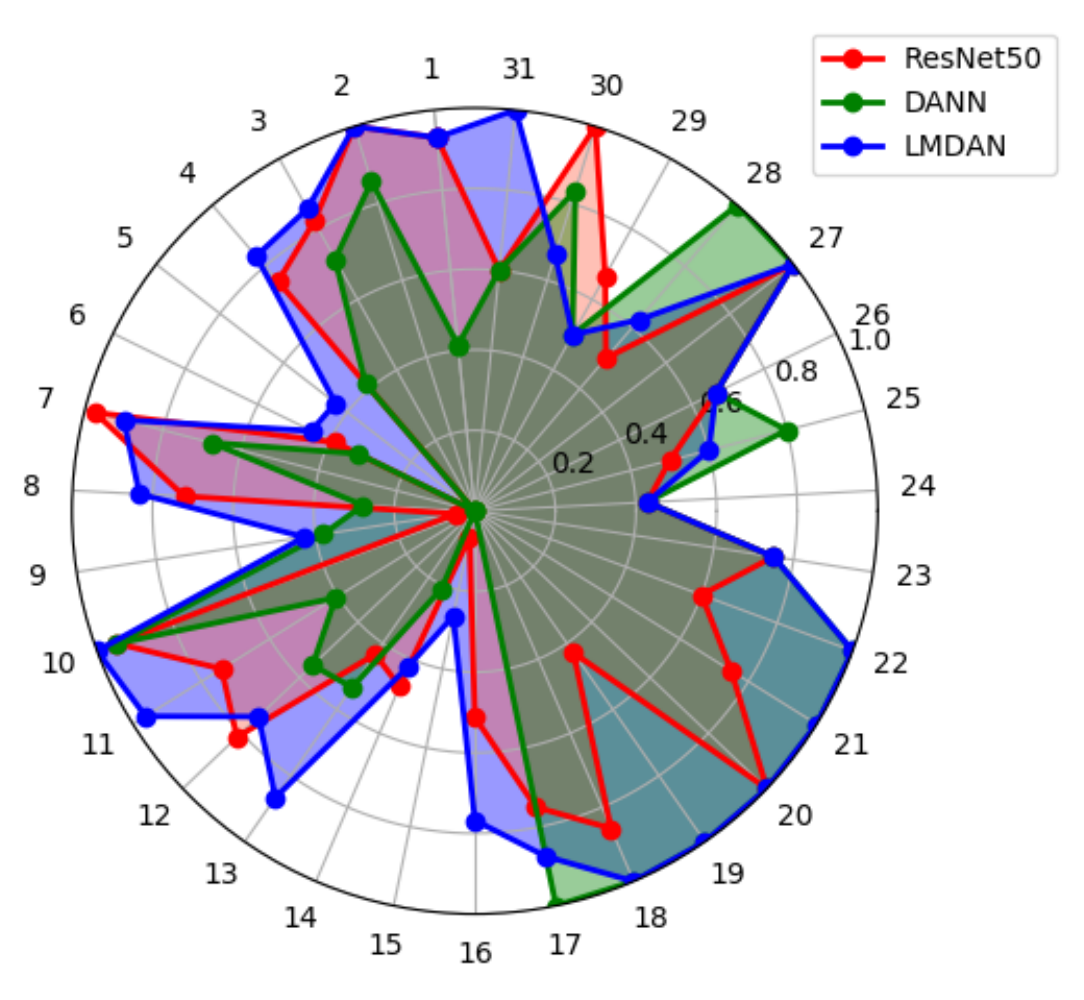}
    \caption{\scriptsize{Class-wise accuracy.}}
    \label{class}
\end{subfigure}
\begin{subfigure}{0.32\columnwidth}
    \includegraphics[width=1.0\columnwidth]{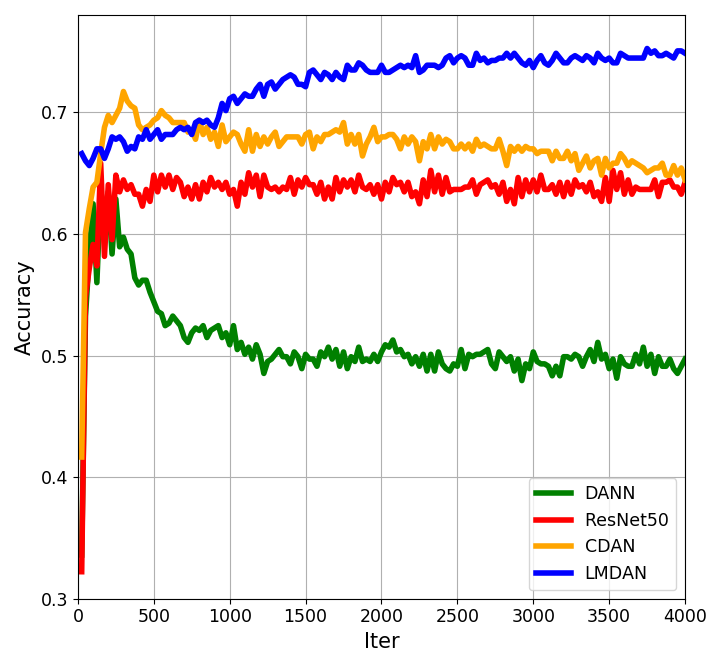}
    \caption{\scriptsize{Training performance.}}
    \label{convergence}
\end{subfigure}
\begin{subfigure}{0.32\columnwidth}
    \includegraphics[width=1.0\columnwidth]{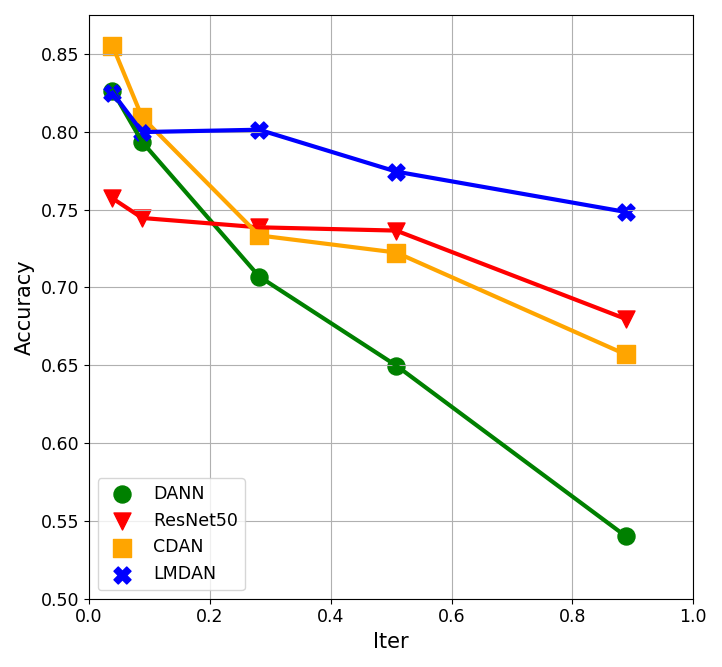}
    \caption{\scriptsize{Different divergences.}}
    \label{divergence}
\end{subfigure}
\caption{\small{Performance of different unsupervised domain adaptation methods on \textbf{A}$\rightarrow$\textbf{W}. (a) shows class-wise predicted accuracy on 31 classes, where the first 15 classes have fewer source samples than the last 16 classes, (b) demonstrates the variation of overall accuracy during training, and (c) reports the performance with different levels of label distribution drift including the original dataset, [0.25;0.25], [0.5;0.5], [0.625;0.625] and [0.75;0.75]. The x-axis denotes the KL-divergence values between source and target label distribution corresponding to 0.0390, 0.0882, 0.2817, 0.5085, and 0.8879.}}
\label{fig:a2w}
\end{figure*}

\section{Experimental Results}

In~\cref{fig:a2w}, we show more experimental details on \textbf{A}$\rightarrow$\textbf{W}. The undesirable performance of other methods mainly results from categories with large sizes on target domain. It is shown that DANN and ResNet-50 return almost 0 accuracy on Class 5\&15. Thanks to label distribution matching, LMDAN achieves much better predictions on these categories. Moreover,~\cref{convergence} shows an increasing performance of LMDAN through iterations, indicating weighting source-target pairs by mini-batch gradually narrows the gap in label space. It is expected to see that all methods degrade with an increasing label distribution divergence in~\cref{divergence}, which demonstrates the divergence in label space has a huge impact on the adaptation performance. LMDAN delivers more robust results than others even under a huge label distribution gap, which is essential in practice due to the agnostic target label distribution. The results~\cref{tab:visda} provide more promising results of LMDAN on \textit{ImageCLEF-DA}[0.75;0.75] and \textit{VisDA-2017}[0.75;0.75]. Our LMDAN shares the same feature alignment component with DANN, and outperforms it by 1.5\% on the original non-modified setting of \textit{VisDA-2017}, which demonstrates the effectiveness of the designated weighting strategy. Note that the dropping ratio is set as 0.75 through general experiments, but similar trends like in~\cref{divergence} toward different dropping ratios are observed, so we do not formally present superfluous experimental results.

\section{Visualization}

\begin{figure*}[t]
\centering 
\begin{subfigure}{0.48\columnwidth}
    \includegraphics[width=0.95\columnwidth]{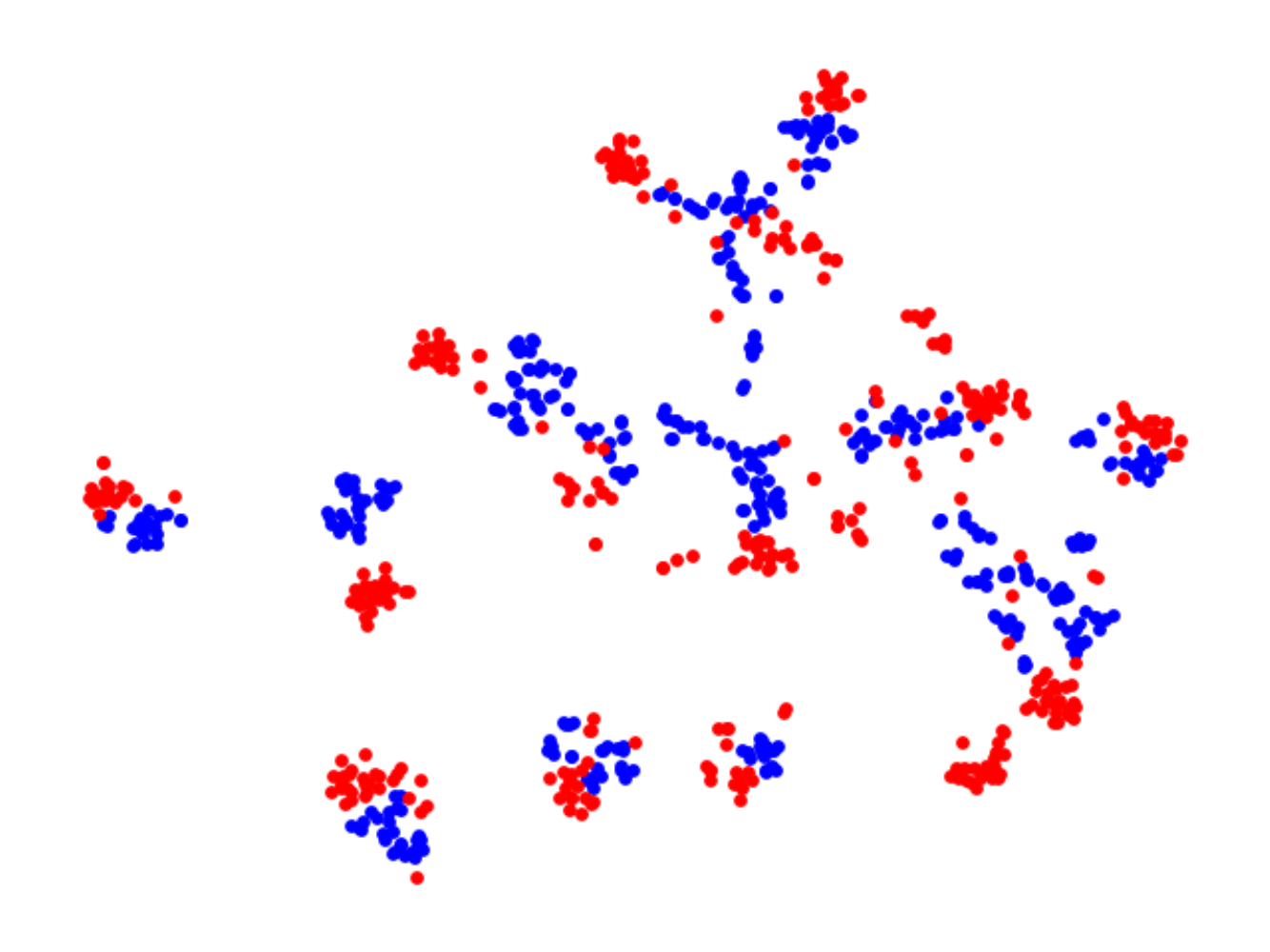}
    \caption{\scriptsize{ResNet50}}
    \label{fig:tsne_resnet}
\end{subfigure}
\begin{subfigure}{0.48\columnwidth}
    \includegraphics[width=0.95\columnwidth]{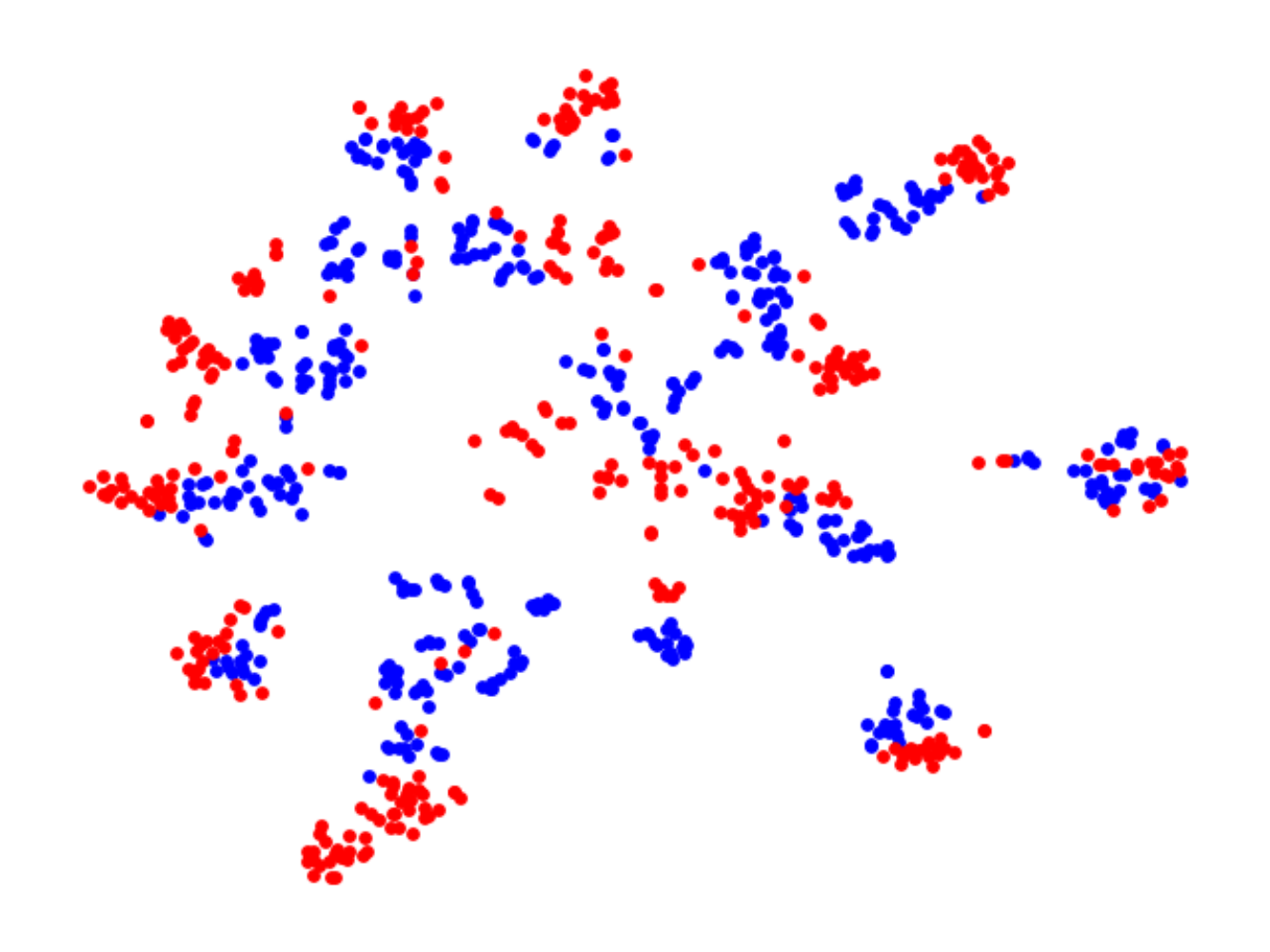}
    \caption{\scriptsize{DANN}}
    \label{fig:tsne_dann}
\end{subfigure}
\begin{subfigure}{0.48\columnwidth}
    \includegraphics[width=0.95\columnwidth]{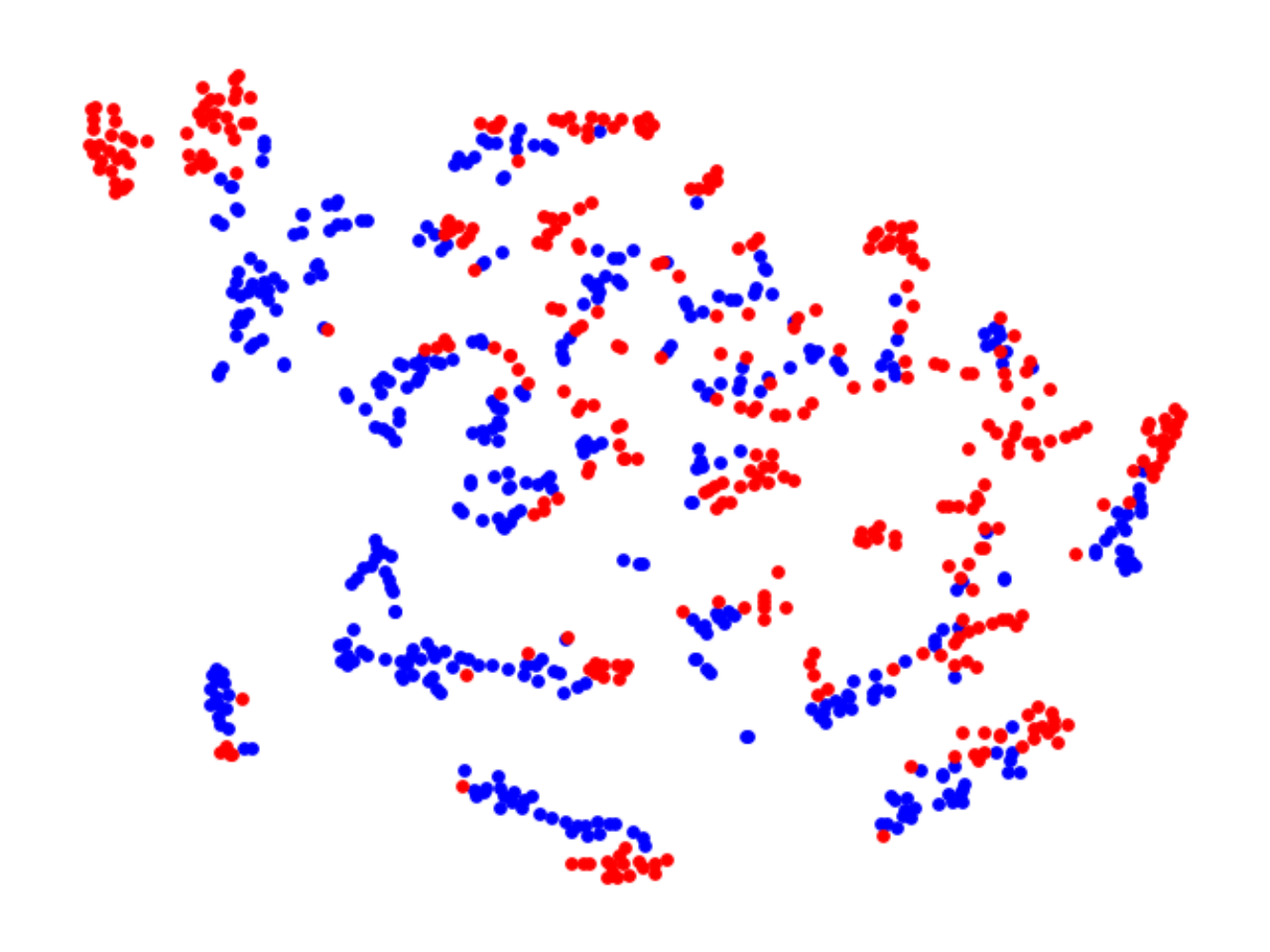}
    \caption{\scriptsize{CDAN}}
    \label{fig:tsne_cdan}
\end{subfigure}
\begin{subfigure}{0.48\columnwidth}
    \includegraphics[width=0.95\columnwidth]{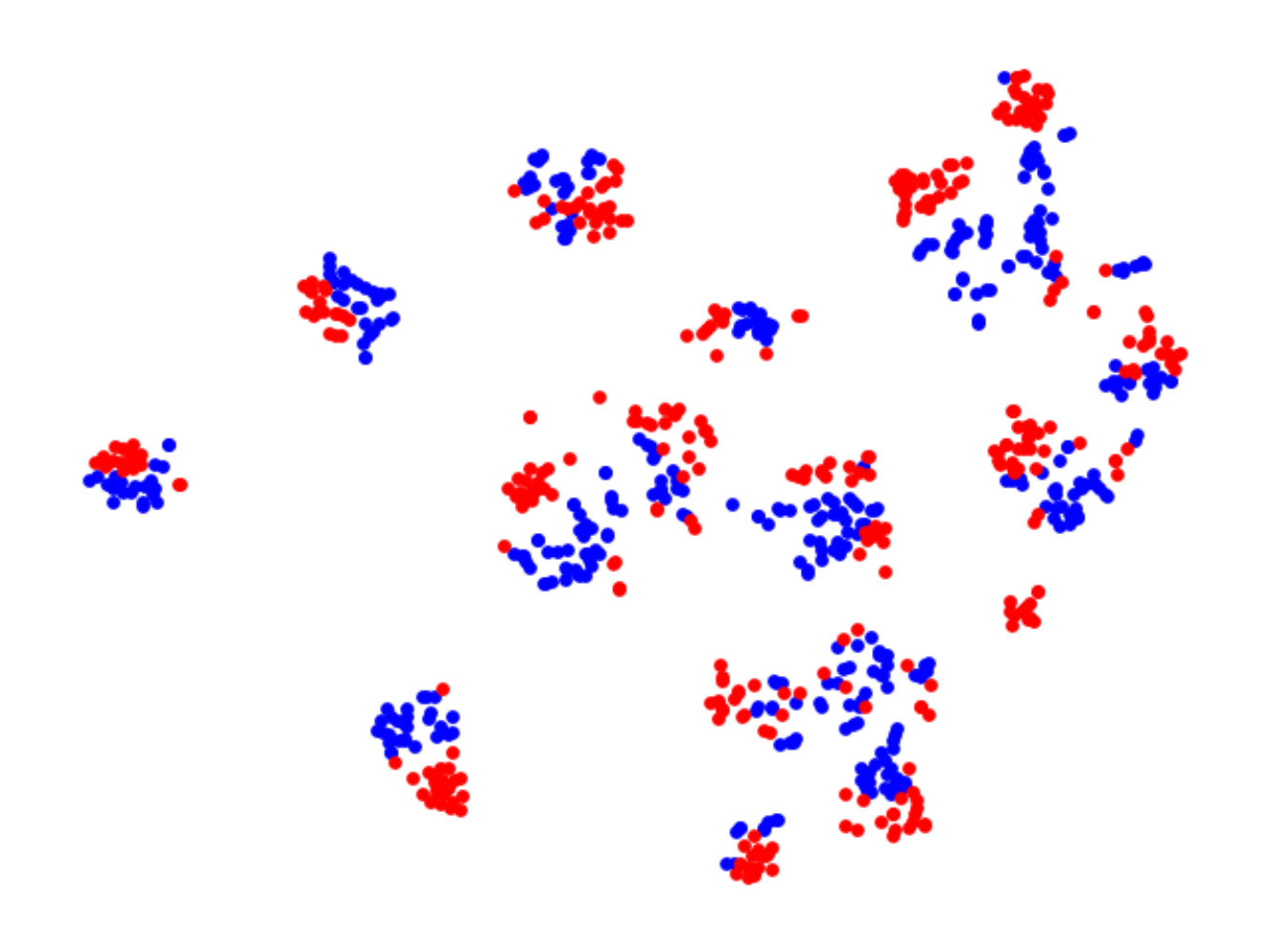}
    \caption{\scriptsize{LMDAN}}
    \label{fig:tsne_ltdan}
\end{subfigure}
\caption{\small{Feature space visualization on \textbf{A}$\rightarrow$\textbf{W}. Red and blue dots denote source and target samples, respectively.}}\label{fig:tsne}
\vspace{-4mm}
\end{figure*}

\cref{fig:tsne} provides embedded feature space visualization results on \textbf{A}$\rightarrow$\textbf{W} on \textit{Office-31}[0.75;0.75] by t-SNE~\cite{tsne}. With huge label distribution divergence between source and target domain, DANN and CDAN cannot preserve the original categorical source structure due to the corruption of representations by aligning mismatched features, and at the same time affect source-target feature alignment dramatically. The negative effects of adaptation further lead to inferior predicted performance on target data. ResNet-50 preserves better source structural features but with less adaptation on domain divergence. LMDAN re-weights samples in source domain, which not only refines the classified decision boundary but also provides better-aligned features.

\begin{figure}[t]
\includegraphics[width=0.95\columnwidth]{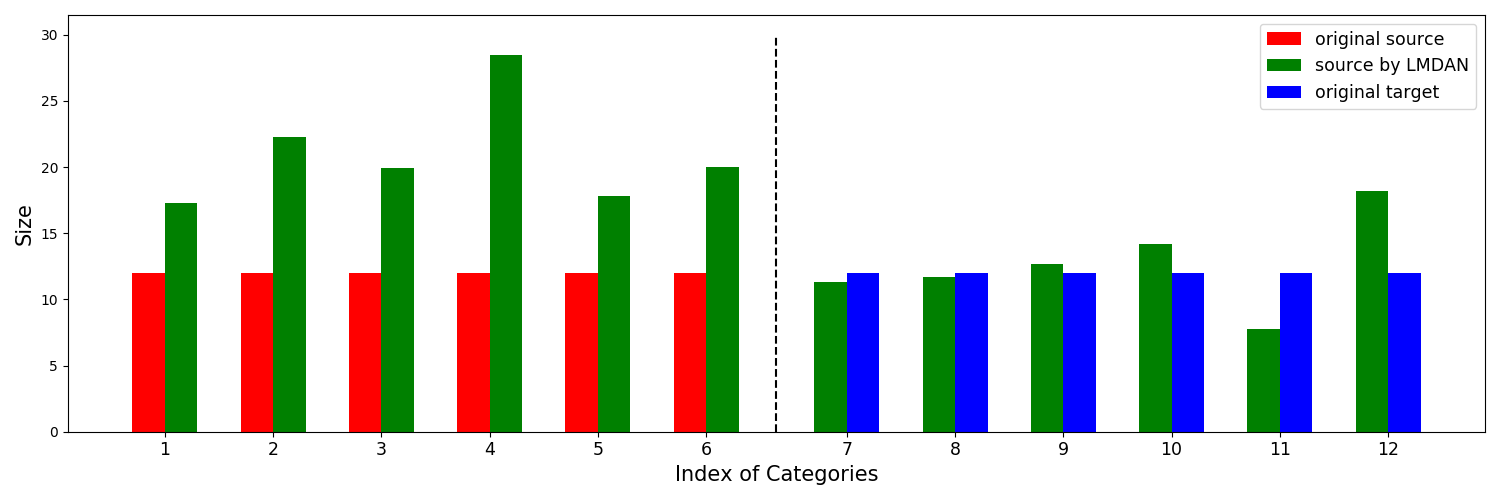}
\caption{Label matching on \textbf{C}$\rightarrow$\textbf{I} of \textit{ImageCLEF-DA}[0.75;0.75], where LMDAN enlarges the first six and shrinks the last six categories of source data.}\label{fig:labeltrans}
\end{figure}

\section{Distribution Matching}

\cref{fig:labeltrans} shows the effectiveness of LMDAN on label distribution matching. \textit{ImageCLEF-DA}[0.75;0.75] has few source samples in the first six categories but more source samples in the rest categories, and target samples work inversely. With label distribution matching in LMDAN, sizes of the first six categories in source domain are assigned with larger weights, and this is equal to enlarging the class size. On the contrary, the rest classes are shrunk to match the target label distribution. Therefore, the source label distribution after matching becomes similar to the target one. With matched distributions between source and target domain in both feature and label space, LMDAN delivers the positive transfer consistently.

\section{Hyperparameter Analysis}

LMDAN employs $\alpha$ as a hyper-parameter to balance the imbalance of the source label distribution and source-target label distribution drift. \cref{tab:hyper} shows the performance of LMDAN on \textbf{D}$\rightarrow$\textbf{A} on modified \textit{Office-31} with different $\alpha$. The second and third columns present the performance with only the first/second term in our weighting function. When $\alpha=0$, label distribution matching provides the inferior performance since the poor performance from the imbalanced classifier on target sample predictions. With an increasing $\alpha$, LMDAN gains improvements with joint actions from source label distribution internal balancing and source-target label distribution match. We set $\alpha=2$ as the default.

\begin{table}[t]
  \centering
  \fontsize{9.5}{9.5}\selectfont
  \caption{Analysis on weight learning in LMDAN on \textbf{D}$\rightarrow$\textbf{A}.}\label{tab:hyper}
    \resizebox{0.8\columnwidth}{!}{
    \begin{tabular}{ccccccc}
        \toprule
        Divergence & w/o second term & $\alpha=0$ & $\alpha=1$ & $\alpha=2$ & $\alpha=3$ & $\alpha=4$ \\
        \midrule
        $[0.25;0.25]$ & 59.7 & 53.9 & 62.3 & \textbf{64.8} & 59.5 & 56.2 \\
        $[0.50;0.50]$ & 60.3 & 34.9 & 50.4 & \textbf{62.4} & 58.9 & 51.5 \\
        $[0.75;0.75]$ & 55.2 & 12.2 & 53.0 & \textbf{57.8} & 57.7 & 53.3 \\
        \bottomrule
    \end{tabular}
  }
\end{table}

\end{document}